\documentclass[10pt,twocolumn,letterpaper]{article}

\usepackage[pagenumbers]{cvpr} 

\definecolor{cvprblue}{rgb}{0.21,0.49,0.74}
\usepackage[pagebackref,breaklinks,colorlinks,allcolors=cvprblue]{hyperref}
\usepackage[accsupp]{axessibility} 
\usepackage{minitoc}
\usepackage{titletoc}
\usepackage{booktabs}
\usepackage{hyperref}
\usepackage{url}
\usepackage{amsmath,amssymb,amsfonts}
\usepackage{mathtools}
\usepackage{graphicx}
\usepackage{textcomp}
\usepackage{booktabs}
\usepackage{xcolor}
\usepackage{algorithmic}
\usepackage[ruled,vlined]{algorithm2e}
\usepackage{multirow}
\usepackage{threeparttable}
\usepackage{array}
\usepackage{float}
\usepackage{wrapfig}
\usepackage{caption}
\usepackage{enumitem}
\usepackage{placeins}
\usepackage[utf8]{inputenc}
\newcommand{\stdsize}[1]{\scalebox{0.85}{#1}}
\newcommand{\appgreen}[1]{\textcolor[rgb]{0.,0.412,0.243}{#1}}
\definecolor{commentcolor}{RGB}{156,156,156}
\newcommand{\comment}[1]{\textcolor{commentcolor}{\# #1}}

\def\paperID{****} 
\def\confName{CVPR}
\def\confYear{2026}

\title{Learning Spatial-Preserving Hierarchical Representations for Digital Pathology}

\author{
Weiyi Wu$^{1}$, Xingjian Diao$^{1}$,  Chunhui Zhang$^{1}$, Chongyang Gao$^{3}$, \\ 
Xinwen Xu$^{2}$, Siting Li$^{1}$, and Jiang Gui$^{1}$\\
$^{1}$Dartmouth College; $^{2}$Massachusetts General Hospital; $^{3}$Northwestern University\\
{\tt\small \{weiyi.wu.gr, xingjian.diao.gr, siting.li, jiang.gui\}@dartmouth.edu} \\ 
{\tt\small xixu3@mgh.harvard.edu,}  \\
{\tt\small chongyanggao2026@u.northwestern.edu}
}

\begin{document}
\maketitle
\begin{abstract}
Whole slide images (WSIs) pose fundamental computational challenges due to their gigapixel resolution and the sparse distribution of informative regions. Existing approaches often treat image patches independently or reshape them in ways that distort spatial context, thereby obscuring the hierarchical pyramid representations intrinsic to WSIs. We introduce Sparse Pyramid Attention Networks (\text{SPAN}), a hierarchical framework that preserves spatial relationships while allocating computation to informative regions. \text{SPAN} constructs multi-scale representations directly from single-scale inputs, enabling precise hierarchical modeling of WSI data. We demonstrate \text{SPAN}'s versatility through two variants: \text{SPAN}-MIL for slide classification and \text{SPAN}-UNet for segmentation. Comprehensive evaluations across multiple public datasets show that \text{SPAN} effectively captures hierarchical structure and contextual relationships. Our results provide clear evidence that architectural inductive biases and hierarchical representations enhance both slide-level and patch-level performance. By addressing key computational challenges in WSI analysis, \text{SPAN} provides an effective framework for computational pathology and demonstrates important design principles for large-scale medical image analysis. Code is available at \text{https://github.com/wwyi1828/SPAN}.
\end{abstract}    
\section{Introduction}
\label{sec:introduction}

Whole Slide Images (WSIs) have become indispensable in modern digital pathology. These high-resolution scans, typically derived from Hematoxylin and Eosin (H\&E)-stained tissue samples, allow precise identification of cellular structures and abnormalities. By digitizing histopathological slides, WSIs enable pathologists to analyze tissue samples across multiple scales, ranging from high-level tissue architecture to fine-grained cellular morphology, thereby supporting more accurate and efficient diagnoses. Beyond manual examination, WSIs facilitate computer-aided diagnosis~\citep{campanella2019clinical, abels2019computational} and serve as the foundation for a variety of computational pathology tasks. At the \textit{patch level}, localized problems such as nuclei segmentation~\citep{lou2024structure, lin2024bonus} and tissue classification~\citep{veeling2018rotation,jiang2023hierarchical, wu2023improving} can be effectively addressed using standard computer vision methods, since the scale is manageable and the regions of interest are well defined.

\begin{figure}[t]
\centering
\includegraphics[width=0.48\textwidth]{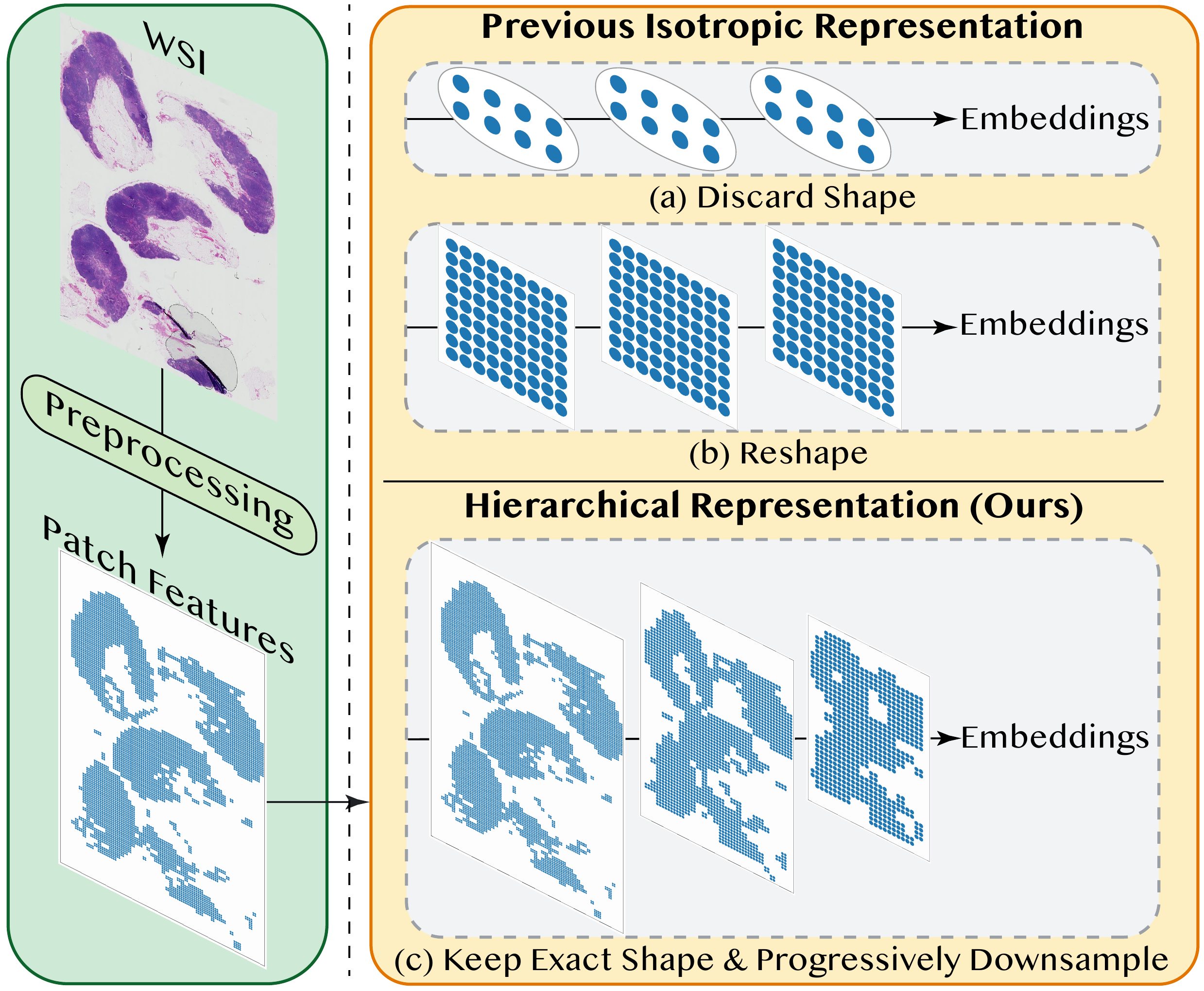}
\caption{Left: A WSI is preprocessed by patch tiling and feature extraction. 
    Right: (a) Patches treated as i.i.d.\ samples. 
    (b) Patches reshaped into squares or flattened. 
    (c) Patches preserved in their original shapes and progressively merged.}
\label{fig:vizabs}
\end{figure}

In contrast, \textit{slide-level} analysis presents fundamentally different computational challenges due to the gigapixel scale of WSIs and the sparse and irregular distribution of informative regions~\citep{lu2021data}. Key slide-level tasks include tumor detection, subtyping, and grading~\citep{brancati2022bracs, bejnordi2017diagnostic, wu2026exploiting, hou2024a}, which rely on histologically grounded labels with relatively low noise. More recently, tasks such as biomarker prediction~\citep{coudray2018classification, jin2024teacher, el2024whole} and survival prediction~\citep{chen2021whole, li2023survival} have drawn increasing interest. Biomarker prediction requires linking visual features to genetic alterations, while survival prediction is often framed as classification via discretized survival times. In these settings, labels are derived from clinical or genomic data and may not correspond directly to visual cues, making the discovery of non-obvious histopathological patterns especially challenging.

Because WSIs often exceed billions of pixels, direct end-to-end analysis is computationally infeasible with conventional vision models. Moreover, large regions of background or non-diagnostic content necessitate preprocessing steps that filter out uninformative patches, resulting in a sparse and irregular distribution of tissue regions across the slide (Fig.\ref{fig:vizabs}). Standard downstream analysis operates on these sparsely distributed patches. A widely adopted strategy treats patches as independent and identically distributed samples~\citep{campanella2019clinical, lu2021data} (Fig.\ref{fig:vizabs}, Top), ignoring spatial relationships entirely. Another line of work reshapes sparse patches by arranging them into dense squares~\citep{shao2021transmil, tang2024feature} or flattening them into sequences so that standard model architectures can be applied (Fig.\ref{fig:vizabs}, Middle). However, such reshaping artificially connects non-adjacent patches, thereby distorting true spatial relationships inherent in the irregular distribution of informative regions. Both strategies either discard or distort the hierarchical spatial organization of WSIs, risking the loss of critical diagnostic information. Our approach instead constructs hierarchical representations that preserve exact spatial relationships and capture multi-scale context (Fig.\ref{fig:vizabs}, Bottom), addressing these limitations.

Transformer models demonstrate remarkable success in modeling long-range dependencies in both language~\citep{devlin2018bert, liu2020roberta} and vision~\citep{dosovitskiy2021an, hatamizadeh2024fastervit, darcet2024vision}. However, applying them directly to WSIs remains infeasible: The quadratic complexity of self-attention is prohibitive at the gigapixel scale~\citep{NIPS2017_3f5ee243}. Although sparse and hierarchical attention variants~\citep{beltagy2020longformer, zaheer2020big, wang2021pyramid, liu2021swin} mitigate this in regularly shaped data, they are poorly suited for WSIs, where informative content is both sparse and irregular. Consequently, WSI-specific Transformer models attempt to circumvent this mismatch by rearranging the irregular spatial distribution of patches. For example, TransMIL~\citep{shao2021transmil} relies on re-squaring the patch layout with Nyström attention and [CLS] tokens, while others introduce region attention after densifying the patch arrangement~\citep{tang2024feature}. These approaches inevitably distort positional information and restrict modeling to isotropic representations, failing to exploit the hierarchical structures that have proven vital in general computer vision.

To address these challenges, we propose the Sparse Pyramid Attention Network (\text{SPAN}), a sparse-native framework for WSI analysis. \text{SPAN} preserves exact spatial information while enabling hierarchical operations such as shifted-window attention and multi-scale feature downsampling, bridging the gap between general computer vision architectures and WSI-specific needs. Its design integrates two complementary modules: the Spatial-Adaptive Feature Condensation (SAC) module, which progressively builds hierarchical representations by condensing informative regions, and the Context-Aware Feature Refinement (CAR) module, which captures complex local and global dependencies at each scale. Together, they direct computation toward diagnostically relevant areas and enable pyramid-style architectures from general vision to be applied to sparse, irregular WSI data. The hierarchical design progressively reduces the number of tokens (approximately 1/4 at each stage), making it more efficient than its non-hierarchical counterpart while maintaining strong modeling capacity.

We validate \text{SPAN} across multiple public datasets~\citep{farahmand2022deep, aresta2019bach, brancati2022bracs, bejnordi2017diagnostic, bandi2018detection} on classification and segmentation tasks. Experiments demonstrate that \text{SPAN} consistently outperforms state-of-the-art methods by capturing spatial and contextual information more effectively. Our main contributions are:

\begin{itemize}[leftmargin=*]
\item A sparse computational framework that preserves spatial relationships in WSIs, enabling the direct use of hierarchical vision techniques.
\item The \text{SPAN} architecture with SAC and CAR modules, which jointly build multi-scale representations through spatial-adaptive condensation and contextual refinement, supporting flexible task-specific variants.
\item Comprehensive evaluations demonstrate that embedding \textit{hierarchical and sparsity-aware inductive biases} into the architecture substantially enhances the representation learning on gigapixel histopathological images.
\end{itemize}
\section{Related Works}
\label{sec:preliminary}

\begin{figure*}[t]
  \centering
  \includegraphics[width=1\linewidth]{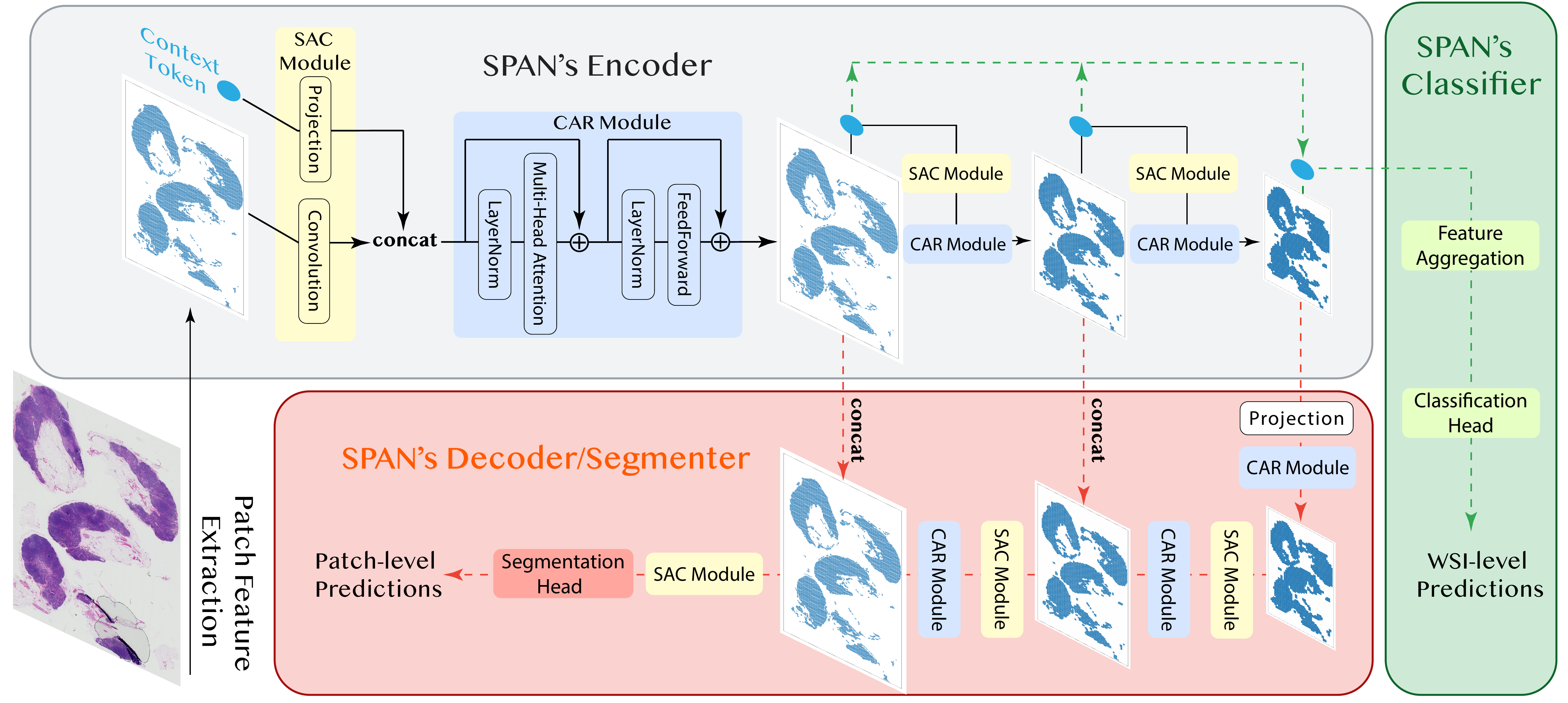}
  \caption{Overall architecture of \text{SPAN}. The encoder begins with a SAC module comprising Projection and Convolution components, followed by CAR that employs window attention through LayerNorm, Multi-Head Attention, and Feed-Forward layers for local context modeling. While the initial SAC preserves spatial dimensions with $1\times 1$ convolution, subsequent SAC modules progressively downsample tokens to approximately 1/4 of their previous token count. This SAC-CAR sequence repeats multiple times for hierarchical feature extraction and refinement. Task-specific paths (dashed lines) enable flexible downstream applications: the decoder/segmenter path utilizes alternating CAR-SAC modules with transposed convolutions in SAC for upsampling and patch-level predictions, while the classifier path employs feature aggregation for WSI-level predictions.}
  \label{fig:pipeline}
  \vspace{-4mm}
\end{figure*}

\subsection{Vision Model Architectures}
\textbf{Self-attention.} The Vision Transformer (ViT)~\citep{dosovitskiy2021an} successfully adapted self-attention mechanisms~\citep{devlin2018bert, brown2020language} for image recognition. However, its quadratic computational complexity is prohibitive for the tens of thousands of patches generated from a single gigapixel WSI. Subsequent work introduced more efficient variants to handle long sequences. These include models with sparse attention patterns like Longformer~\citep{beltagy2020longformer} and BigBird~\citep{zaheer2020big}, and models with window attention like the Swin Transformer~\citep{liu2021swin}. By computing attention locally within windows and building a hierarchical representation, Swin Transformer achieves linear complexity and captures multi-scale features, leading to state-of-the-art performance on many vision tasks.

Despite these advancements, a fundamental challenge remains in applying these mechanisms to WSIs. They are designed for dense, continuously distributed data. In contrast, the informative patches in WSIs are sparsely and irregularly distributed across a vast, uninformative background. This mismatch makes it inherently difficult to directly apply window-based or dense-matrix-based sparse attention techniques, necessitating specialized approaches that can natively handle sparse data distributions.

\noindent\textbf{Pyramid Structures.} Multi-scale feature representation is a cornerstone of modern computer vision. In CNNs, this is achieved through progressive downsampling~\citep{he2016deep} and explicit pyramid architectures that capture context at multiple resolutions, such as SPP-Net~\citep{he2015spatial}, FPN~\citep{lin2017feature}, and HRNet~\citep{wang2020deep}. This powerful paradigm is successfully integrated into vision transformers as well. Models like Pyramid Vision Transformer (PVT)~\citep{wang2021pyramid} and Swin Transformer~\citep{liu2021swin} incorporate hierarchical designs with efficient attention, proving the value of multi-scale learning.

However, these successful pyramid structures are all designed for dense and uniformly distributed data. They rely on regular downsampling operations (e.g., strided convolutions or patch merging) that are fundamentally inappropriate for the sparse and irregular spatial layout of WSIs. The unique challenges posed by vast uninformative regions prevent the direct application of general-purpose pyramid architectures, leaving a critical gap in WSI analysis.

\subsection{Methods for WSIs}

\textbf{Isotropic Paradigms.} WSIs inherently possess a hierarchical structure, enabling pathologists to examine tissue samples across multiple magnification levels. This multi-scale nature of WSIs underscores the importance of capturing and integrating information from different scales for accurate analysis. However, most existing computational methods fail to fully exploit this characteristic, operating in an isotropic manner—maintaining constant spatial resolution and feature dimensions throughout processing, without the hierarchical downsampling that enables efficient multi-scale reasoning. Mainstream WSI analysis techniques treat patches as independent and identically distributed (i.i.d.) samples, completely disregarding spatial relationships~\citep{ilse2018attention, lu2021data, li2021dual, zhang2022dtfd, tang2023multiple}. Attention-based Multiple Instance Learning (ABMIL)~\citep{ilse2018attention} serves as a foundational approach, aggregating patch-level features for slide-level prediction. Extensions like CLAM~\citep{lu2021data} and DTFD-MIL~\citep{zhang2022dtfd} introduce additional losses or training strategies but still neglect spatial context.

Even methods that attempt to incorporate spatial information remain fundamentally isotropic while introducing additional distortions. TransMIL and its variants~\citep{shao2021transmil,tang2024feature} reshape sparse patches into dense 2D grids, while other approaches~\citep{yang2024mambamil, zheng2025m3amba, fillioux2023structured} flatten patches into sequences. Both strategies forcibly convert sparse inputs into dense representations, also distorting real positional relationships by artificially connecting non-adjacent patches. Crucially, all these approaches process patches at uniform resolution with fixed feature dimensions throughout the network, failing to leverage hierarchical modeling capabilities that have proven crucial in general computer vision tasks. Consequently, WSI analysis has been unable to benefit from key technical advances that have revolutionized general visual tasks.

\noindent\textbf{Hierarchical Paradigms.} Extending SPAN to support multi-scale or multi-resolution inputs is natural. Its hierarchical sparse design may offer a more coherent way to integrate information across magnifications than existing isotropic multi-scale approaches, including HIPT~\citep{chen2022scaling}, H2MIL~\citep{hou2022h}, and ZoomMIL~\citep{thandiackal2022differentiable}. However, these approaches do not build a feature pyramid organically from a single-scale input as in general computer vision. Instead, they depend on multi-scale inputs, requiring the system to process separate patches from multiple magnification levels (e.g., 5x, 10x, 20x). This strategy introduces significant computational and data management overhead. More importantly, within each scale, these methods still operate isotropically, failing to form a cohesive, end-to-end hierarchical representation. This architectural compromise means the central challenge of building a true feature pyramid from a single-scale input remains largely unaddressed. As a result, WSI analysis has yet to fully harness the powerful hierarchical architectures that are now leading in the broader vision community.

\section{Method}
\label{sec:method}

The core of our backbone is a rulebook-based mechanism: a pre-computed set of instructions that explicitly defines input-output mappings for sparse data. This allows for highly efficient computation by targeting only active features and eliminating redundant operations on empty regions. The \text{SPAN} backbone is constructed from a repeating sequence of SAC and CAR modules that adhere to this principle. As illustrated in Fig. \ref{fig:pipeline}, the SAC module performs spatial condensation and coarse-grained feature transformation, while the subsequent CAR module employs transformer blocks with shifted windows for fine-grained contextual refinement. This complementary design allows the \text{SPAN} backbone to efficiently capture both multi-scale patterns and their long-range dependencies, which can then be utilized by task-specific variants: \text{SPAN}-MIL for classification through global token aggregation, and \text{SPAN}-UNet for segmentation through hierarchical decoding.

This hierarchical processing repeats with subsequent SAC-CAR modules operating on increasingly condensed features, enabling \text{SPAN} to learn pyramid representations that unify multi-granularity information with global understanding. The gradual reduction in spatial resolution  allows \text{SPAN} to efficiently manage memory consumption at deeper layers while preserving multi-scale diagnostic patterns.

\subsection{Spatial-Adaptive Feature Condensation}
The SAC module progressively condenses patches into more compact representations through learnable feature transformations. The design of SAC is motivated by two key insights: the inherent multi-scale nature of histopathological diagnosis that pathologists perform, and the computational efficiency required for processing large-scale WSIs. This motivates us to design an adaptive feature extraction that can handle the irregular spatial distribution of patches.

Our condensation process maintains spatial relationships while progressively reducing spatial dimensions to capture multi-scale patterns. To achieve this efficiently, we implement SAC using sparse convolutions~\citep{liu2015sparse} for downsampling and hierarchical feature encoding. This choice naturally aligns with the WSI structure, where significant background portions contain uninformative regions, enabling selective computation only where meaningful features are present.

\noindent\textbf{Sparse Convolution Rulebook.} The key challenge in processing sparse WSI data is to perform convolution operations efficiently without densifying the entire spatial grid. Our solution is a rulebook-based mechanism that precomputes which spatial locations interact during convolution, enabling targeted computation only at active positions. Conceptually, the rulebook serves as a lookup table: given input coordinates and convolution parameters (kernel size, stride, dilation), it determines the precise input-output mappings required for each convolution operation. This approach avoids processing empty background regions while preserving exact spatial relationships.

Formally, sparse convolution operations manage computation through structured indexing. An index matrix \(\mathbf{I} = \begin{bmatrix} 1 & 2 & \cdots & N \end{bmatrix}^\text{T}\) corresponds to the coordinate matrix \(\mathbf{P} = [ p_{i} \mid i \in \mathbf{I} ] \in \mathbb{N}^{N \times 2}\) and the feature matrix \(\mathbf{X} = [ x_{i} \mid i \in \mathbf{I} ] \in \mathbb{R}^{N \times d}\). This structured representation ensures efficient access to coordinates and their associated features during sparse convolution operations.

For each convolutional layer, the output coordinates are computed based on the input coordinates, the kernel size \( K \), the dilation \( D \), and the layer's stride \( S \):
\begin{equation}
\label{eq:output_coordinates}
\mathbf{P}_{\text{out}} = \{ p_{i_{\text{out}}} \mid p_{i_{\text{out}}} = \left\lfloor \frac{p_{i_{\text{in}}} - (K-1) \cdot D}{S} \right\rfloor,\ \forall p_{i_{\text{in}}} \in \mathbf{P}_{\text{in}} \},
\end{equation}
where \( \left\lfloor \cdot \right\rfloor \) denotes the floor operation, and \( (K-1) \cdot D \) adjusts for the expansion of the receptive field due to the kernel size and dilation. The corresponding output indices \( \mathbf{I}_{\text{out}} \) are assigned sequentially starting from 1.

To determine the valid mappings between input and output indices for each kernel offset, we construct a \textit{rulebook} \( \mathcal{R}_{k} \) defined as:
\begin{equation}
\label{eq:rulebook}  
\mathcal{R}_{k} = \left\{ (i_{\text{in}}, i_{\text{out}}) \mid p_{i_{\text{in}}} + k = p_{i_{\text{out}}} \right\}, \quad k \in \mathcal{K},
\end{equation}
where \( \mathcal{K} \) is the set of kernel offsets, and \( p_{i_{\text{in}}} \) and \( p_{i_{\text{out}}} \) are input and output coordinates, respectively. Each entry in \( \mathcal{R}_{k} \) represents an atomic operation, specifying that the input position \( p_{i_{\text{in}}} \) shifted by the kernel offset \( k \) matches the output position \( p_{i_{\text{out}}} \). The complete rulebook \( \mathcal{R}_\mathcal{K} = \bigcup_{k \in \mathcal{K}} \mathcal{R}_{k} \) efficiently encodes the locations and conditions under which convolution operations are to be performed.

Each sparse convolutional layer performs convolution by executing the atomic operations defined in the rulebook \(\mathcal{R}_\mathcal{K}\). An atomic operation \((i_{\text{in}}, i_{\text{out}}) \in \mathcal{R}_k\) transforms the input feature \(h_{i_{\text{in}}}\) using the corresponding weight matrix \(W_l(k)\) and accumulates the result to the output feature \(h_{i_{\text{out}}}\). The complete sparse convolution operation for a layer \(l\) is defined as:
\begin{equation}
\label{eq:sparse_conv_general_updated}
h_{i_{\text{out}}} = \sum_{k \in \mathcal{K}} \sum_{\mathcal{R}_k} W_l(k) h_{i_{\text{in}}} + b_l,
\end{equation}
where \( h_{i_{\text{in}}} \in \mathbb{R}^{d_{\text{in}}} \) is the input feature at index \( i_{\text{in}} \), \( h_{i_{\text{out}}} \in \mathbb{R}^{d_{\text{out}}} \) is the output feature at index \( i_{\text{out}} \), \( W_l(k) \in \mathbb{R}^{d_{\text{out}} \times d_{\text{in}}} \) is the weight matrix associated with kernel offset \( k \), and \( b_l \in \mathbb{R}^{d_{\text{out}}} \) is the bias term for layer \( l \).

Using this rulebook-based approach, the sparse convolutional layer efficiently aggregates information from neighboring input features by performing computations only at the necessary locations. This method effectively captures local spatial patterns in the sparse data while significantly reducing computational overhead and memory usage compared to dense convolution operations, as it avoids unnecessary calculations in empty or uninformative regions. For the context token, we compute and average features with all kernel weights and biases if dimension reduction is needed. Otherwise, we maintain an identity projection.

\begin{figure}[ht]
\centering
\includegraphics[width=0.47\textwidth]{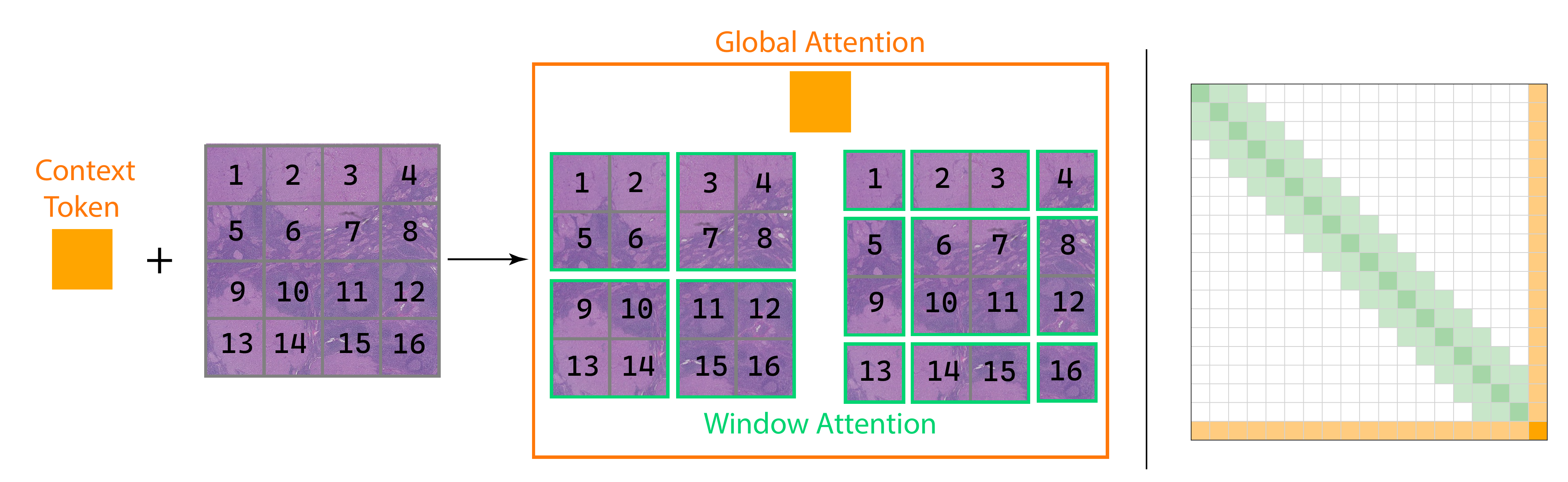}
\caption{Schematic of CAR. \textit{Left:} The input is partitioned into overlapping $2w \times 2w$ windows. Attention is computed locally within windows (green box) and globally via a learnable token that attends to all tokens in the input sequence (orange box). \textit{Right:} The attention matrix visualizes this: diagonal blocks (green) show local attention, while the full row/column (orange) highlights the global token’s unrestricted scope across all positions.}
\vspace{-2mm}
\label{fig:attention}
\vspace{-4mm}
\end{figure}

\subsection{Context-Aware Feature Refinement}
The CAR module builds upon the condensed feature representation to model comprehensive contextual relationships. While the preceding SAC module efficiently captures hierarchical features through progressive condensation, the refined understanding of histological patterns requires modeling both local tissue structures and their long-range dependencies. This dual modeling requirement motivates us to adopt attention mechanisms, which excel at capturing both local and long-range dependencies through learnable interactions between features.

To effectively implement the CAR module, we face several technical challenges in applying attention mechanisms to WSI analysis. Traditional sparse attention approaches~\citep{liu2021swin, beltagy2020longformer, zaheer2020big}, despite their success in various domains, operate on dense feature matrices by striding over fixed elements in the matrix's memory layout. This approach requires densifying our sparse WSI features and applying padding operations to match the fixed memory layout. Given the high feature dimensionality characteristic of WSI analysis, such transformation would introduce substantial memory and computational overhead while compromising the efficiency established in the previous SAC module. Therefore, we develop a sparse attention rulebook that directly operates on the sparse feature representation, maintaining compatibility with the SAC module's index-coordinate system. Our approach leverages $\mathbf{I}$ and $\mathbf{P}$ inherited from previous layers to define sparse attention windows, where features within each window can attend to each other without dense transformations. This design preserves both computational efficiency and the sparse structure compatibility established in earlier modules.

\noindent\textbf{Sparse Attention Rulebook.} To efficiently handle sparse data representations, we formulate attention computation using rulebooks following the paradigm of sparse convolutions. The first step is to generate attention windows that define which tokens should attend to each other. For efficient window generation, we temporarily densify $\mathbf{I} \in \mathbb{N}^{N}$ into a regular grid using patch coordinates $\mathbf{P} \in \mathbb{N}^{N \times 2}$ with zero padding. This enables efficient block-wise memory access on a low-dimensional index matrix rather than operating on a high-dimensional feature matrix. As illustrated in Fig.~\ref{fig:attention}, we stride over the densified index matrix to generate regular and shifted windows, where the shifting operation ensures comprehensive coverage of local contexts. The resulting $\mathcal{W}$ is a collection of windows, where each window contains a set of patch indices excluding padded zeros. These windows effectively define the grouping of indices for constructing an attention rulebook.

To enhance the model's ability to capture global dependencies, we introduce a learnable global context token that provides a shared context accessible to all other tokens. The combined hidden features can be represented as $\mathbf{H} = [h_{i_1}^\top, h_{i_2}^\top, \ldots, h_{i_N}^\top, h_g^\top] \in \mathbb{R}^{(N + 1) \times d_\text{out}}$, where $h_g$ denotes the global context token. For self-attention computation, we project $\mathbf{H} \in \mathbb{R}^{(N+1) \times d}$ into $\mathbf{Q}$, $\mathbf{K}$, and $\mathbf{V}$ using linear projections.

Having defined the attention windows, we now construct two types of rulebooks to capture both local and global dependencies. For local attention, the rulebook $\mathcal{R}_w$ for each window is defined as:

\begin{equation}
\mathcal{R}_w = \left\{ (i, j) \mid i, j \in w \right\},\quad w \in \mathcal{W},
\end{equation}
where $\mathcal{W}$ denotes the set of all attention windows, and $i$ and $j$ represent the indices of the input and output patches within the window $w$, respectively. Each entry $(i, j) \in \mathcal{R}_w$ represents a local attention atomic operation between tokens $i$ and $j$. These atomic operations are defined by the following equations. The attention scores are computed with learnable positional bias to account for spatial relationships:
\begin{equation}
\label{eq:local_attn}
e_{ij}^{\text{local}} = \frac{\mathbf{q}_i^\top \mathbf{k}_j}{\sqrt{d}} + B(p_i - p_j),
\end{equation}
where $\mathbf{q}_i$ and $\mathbf{k}_j$ represent the query and key vectors for local tokens $i$ and $j$, respectively, and $p_i$ and $p_j$ denote their positions. $B(p_i - p_j)$ represents the learnable relative positional biases (RPB)~\citep{liu2021swin}, parameterized by a matrix $B \in \mathbb{R}^{(2w_{size}-1) \times (2w_{size}-1) \times \text{num\_heads}}$.

The choice of positional encoding is crucial for capturing spatial relationships in WSI analysis. RPB enhances the model's ability to recognize positional nuances and disrupt the permutation invariance inherent in self-attention mechanisms while maintaining parameter efficiency. Alternative approaches present different trade-offs: absolute positional encoding (APE)~\citep{dosovitskiy2021an} would significantly increase the parameter count given the extensive spatial dimension of possible positions in WSIs, while Rotary Position Embedding (RoPE)~\citep{heo2024ropevit,su2024roformer} and Attention with Linear Biases (Alibi)~\citep{press2022train}, despite their parameter efficiency in language models, prove less effective at capturing spatial relationships in our context.

The final output of the local attention is computed as:
\begin{equation}
\label{eq:sum_local}
\mathbf{h}_i^{\text{local}} = \sum_{w \in \mathcal{W}} \sum_{j: (i, j) \in \mathcal{R}_{\text{w}}} \frac{\exp(e_{ij}^{\text{local}})}{\sum_{k: (i, k) \in \mathcal{R}_{\text{local}}} \exp(e_{ik}^{\text{local}})} \mathbf{v}_j.
\end{equation}

To complement local attention with global context modeling, we introduce global attention that operates on all patch tokens and the learnable global context token. The global attention rulebook is defined as:
\begin{equation}
\mathcal{R}_g = \left\{ (i, j), (j, i) \mid i \in [1,N], j \in \{N+1\} \right\}.
\end{equation}

The global attention mechanism employs similar formulations as equations \eqref{eq:local_attn} and \eqref{eq:sum_local} but excludes the positional bias term, yielding $\mathbf{h}_i^{global}$. While local attention is constrained to windows, global attention spans across the entire feature map through the global context token, enabling comprehensive contextual integration. The final output features combine both local and global dependencies through:
\begin{equation}
\mathbf{h}_i^{\text{out}} = \mathbf{h}_i^{local} + \mathbf{h}_i^{global}.
\end{equation}

For downstream tasks, \text{SPAN} serves s a backbone that support task-specific variants: \textbf{\text{SPAN}-MIL} employs global token aggregation for slide-level classification tasks, while \textbf{\text{SPAN}-UNet} utilizes a U-Net-style decoder for patch-level segmentation tasks (Details in Appendix A.1).

\begin{table*}[ht]
    \caption{Classification performance comparison of MIL methods on CAMELYON16, Yale HER2, and BRACS datasets using different feature extractors. {\textcolor{blue}{\textbullet}} ABMIL-based, {\textcolor{purple}{\textbullet}} Transformer-based, {\appgreen{\textbullet}} \text{SPAN}-based.}
    \label{tab:compare_all_datasets}
    \centering
    \resizebox{0.95\linewidth}{!}{
    \begin{tabular}{lcccccc}
       \toprule
       & \multicolumn{2}{c}{\textbf{CAMELYON16}} & \multicolumn{2}{c}{\textbf{Yale HER2}} & \multicolumn{2}{c}{\textbf{BRACS}} \\
       \cmidrule(lr){2-3} \cmidrule(lr){4-5} \cmidrule(lr){6-7}
       Method & Accuracy & F1 & Accuracy & F1 & Accuracy & Macro F1 \\
       \midrule
       \multicolumn{7}{l}{\textbf{General ResNet50 Feature}} \\
       \midrule
       {\textcolor{blue}{\textbullet}} ABMIL    & $0.857 \pm 0.085$ & $0.850 \pm 0.088$ & $0.687 \pm 0.084$ & $0.664 \pm 0.091$ & $0.687 \pm 0.023$ & $0.552 \pm 0.039$ \\
       {\textcolor{blue}{\textbullet}} CLAM-SB  & $0.873 \pm 0.040$ & $0.868 \pm 0.039$ & $\underline{0.713} \pm 0.084$ & $\underline{0.699} \pm 0.090$ & $0.687 \pm 0.044$ & $0.562 \pm 0.041$ \\
       {\textcolor{blue}{\textbullet}} CLAM-MB  & $0.867 \pm 0.031$ & $0.862 \pm 0.031$ & $0.693 \pm 0.089$ & $0.684 \pm 0.094$ & $0.696 \pm 0.039$ & $0.545 \pm 0.049$ \\
       {\textcolor{blue}{\textbullet}} DSMIL    & $0.887 \pm 0.051$ & $0.881 \pm 0.050$ & $0.693 \pm 0.060$ & $0.676 \pm 0.049$ & $0.699 \pm 0.035$ & $0.553 \pm 0.056$ \\
       {\textcolor{blue}{\textbullet}} MHIM     & $0.883 \pm 0.053$ & $0.877 \pm 0.056$ & $0.706 \pm 0.104$ & $0.695 \pm 0.100$ & $0.716 \pm 0.028$ & $0.560 \pm 0.066$ \\
       {\textcolor{blue}{\textbullet}} ACMIL    & $\underline{0.893} \pm 0.015$ & $\underline{0.889} \pm 0.011$ & $\underline{0.713} \pm 0.030$ & $0.685 \pm 0.045$ & $\underline{0.720} \pm 0.022$ & $\underline{0.604} \pm 0.074$ \\
       {\textcolor{purple}{\textbullet}} TransMIL & $0.873 \pm 0.053$ & $0.867 \pm 0.053$ & $0.672 \pm 0.085$ & $0.652 \pm 0.113$ & $0.692 \pm 0.037$ & $0.577 \pm 0.034$ \\
       {\textcolor{purple}{\textbullet}} RRT      & $0.867 \pm 0.029$ & $0.862 \pm 0.027$ & $0.647 \pm 0.069$ & $0.631 \pm 0.072$ & $0.718 \pm 0.036$ & $0.595 \pm 0.065$ \\
       {\text{\textbullet}} \text{SPAN}-MIL & $\textbf{0.903} \pm 0.030$ & $\textbf{0.898} \pm 0.032$ & $\textbf{0.727} \pm 0.072$ & $\textbf{0.720} \pm 0.070$ & $\textbf{0.725} \pm 0.038$ & $\textbf{0.641} \pm 0.076$ \\
       \midrule
       \multicolumn{7}{l}{\textbf{Pathology-specific UNI Feature}} \\
       \midrule
       {\textcolor{blue}{\textbullet}} ABMIL   & $0.980 \pm 0.007$ & $0.978 \pm 0.009$ & $0.813 \pm 0.045$ & $0.804 \pm 0.044$ & $0.761 \pm 0.053$ & $0.667 \pm 0.061$ \\
       {\textcolor{blue}{\textbullet}} CLAM-SB & $\underline{0.990} \pm 0.009$ & $\underline{0.989} \pm 0.010$ & $0.807 \pm 0.028$ & $0.795 \pm 0.031$ & $0.749 \pm 0.062$ & $0.674 \pm 0.047$ \\
       {\textcolor{blue}{\textbullet}} CLAM-MB & $\underline{0.990} \pm 0.015$ & $\underline{0.989} \pm 0.016$ & $0.833 \pm 0.024$ & $0.824 \pm 0.027$ & $0.750 \pm 0.018$ & $0.673 \pm 0.064$ \\
       {\textcolor{blue}{\textbullet}} DSMIL   & $0.983 \pm 0.017$ & $0.982 \pm 0.018$ & $0.827 \pm 0.072$ & $0.820 \pm 0.071$ & $0.764 \pm 0.046$ & $\underline{0.679} \pm 0.009$ \\
       {\textcolor{blue}{\textbullet}} MHIM    & $0.970 \pm 0.025$ & $0.967 \pm 0.026$ & $\underline{0.840} \pm 0.076$ & $\underline{0.833} \pm 0.072$ & $0.766 \pm 0.064$ & $0.677 \pm 0.043$ \\
       {\textcolor{blue}{\textbullet}} ACMIL   & $\underline{0.990} \pm 0.009$ & $\underline{0.989} \pm 0.010$ & $\underline{0.840} \pm 0.015$ & $0.830 \pm 0.025$ & $0.771 \pm 0.019$ & $\underline{0.679} \pm 0.061$ \\
       {\textcolor{purple}{\textbullet}} TransMIL & $0.957 \pm 0.035$ & $0.951 \pm 0.039$ & $0.833 \pm 0.063$ & $\underline{0.833} \pm 0.059$ & $0.740 \pm 0.081$ & $0.649 \pm 0.084$ \\
       {\textcolor{purple}{\textbullet}} RRT     & $0.980 \pm 0.007$ & $0.978 \pm 0.008$ & $0.827 \pm 0.028$ & $0.818 \pm 0.030$ & $\underline{0.776} \pm 0.029$ & $0.672 \pm 0.055$ \\
       {\appgreen{\textbullet}} \text{SPAN}-MIL & $\textbf{0.993} \pm 0.009$ & $\textbf{0.993} \pm 0.010$ & $\textbf{0.860} \pm 0.037$ & $\textbf{0.856} \pm 0.036$ & $\textbf{0.778} \pm 0.028$ & $\textbf{0.690} \pm 0.080$ \\
       \bottomrule
    \end{tabular}
    }
 \end{table*}

\begin{table*}[t]
    \centering
    \caption{Segmentation performance comparison of MIL-based methods on CAMELYON16, SegCAMELYON, Yale HER2, and BACH datasets using different feature extractors.}
    \label{tab:segmentation_by_feature_extractor}
    \resizebox{1\linewidth}{!}{
      \begin{threeparttable}
        \begin{tabular}{lcccccccc}
          \toprule
          Method & \multicolumn{2}{c}{\textbf{CAMELYON16}} & \multicolumn{2}{c}{\textbf{SegCAMELYON}} & \multicolumn{2}{c}{\textbf{Yale HER2}} & \multicolumn{2}{c}{\textbf{BACH}} \\
          \cmidrule(r){2-9}
          & \multicolumn{1}{c}{Dice} & \multicolumn{1}{c}{IoU} & \multicolumn{1}{c}{Dice} & \multicolumn{1}{c}{IoU} & \multicolumn{1}{c}{Dice} & \multicolumn{1}{c}{IoU} & \multicolumn{1}{c}{Dice} & \multicolumn{1}{c}{IoU} \\
          \midrule
          \multicolumn{9}{l}{\textbf{General ResNet50 Feature}} \\
          \midrule
          ABMIL\textsuperscript{\textdagger} & 0.742\stdsize{${\pm}$0.012} & 0.591\stdsize{${\pm}$0.016} & 0.738\stdsize{$\pm$0.038} & 0.586\stdsize{$\pm$0.047} & 0.522\stdsize{${\pm}$0.053} & 0.354\stdsize{${\pm}$0.048} & 0.690\stdsize{${\pm}$0.158} & 0.544\stdsize{${\pm}$0.181} \\
          TransMIL\textsuperscript{\textdagger} & 0.822\stdsize{${\pm}$0.051} & 0.700\stdsize{${\pm}$0.071} & 0.818\stdsize{${\pm}$0.055} & 0.695\stdsize{${\pm}$0.079} & 0.552\stdsize{${\pm}$0.050} & 0.382\stdsize{${\pm}$0.048} & 0.723\stdsize{${\pm}$0.176} & 0.588\stdsize{${\pm}$0.201} \\
          RRT\textsuperscript{\textdagger} & 0.836\stdsize{${\pm}$0.062} & 0.722\stdsize{${\pm}$0.094} & \underline{0.829}\stdsize{${\pm}$0.066} & \underline{0.712}\stdsize{${\pm}$0.100} & 0.546\stdsize{${\pm}$0.050} & 0.377\stdsize{${\pm}$0.048} & 0.705\stdsize{${\pm}$0.128} & 0.557\stdsize{${\pm}$0.159} \\
          GCN & \underline{0.841}\stdsize{${\pm}$0.006} & \underline{0.726}\stdsize{${\pm}$0.010} & 0.809\stdsize{${\pm}$0.068} & 0.684\stdsize{${\pm}$0.098} & 0.555\stdsize{${\pm}$0.050} & 0.386\stdsize{${\pm}$0.048} & 0.695\stdsize{${\pm}$0.169} & 0.552\stdsize{${\pm}$0.191} \\
          GAT & 0.795\stdsize{${\pm}$0.029} & 0.661\stdsize{${\pm}$0.040} & 0.805\stdsize{${\pm}$0.045} & 0.676\stdsize{${\pm}$0.064} & \underline{0.567}\stdsize{${\pm}$0.059} & \underline{0.398}\stdsize{${\pm}$0.058} & 0.715\stdsize{${\pm}$0.136} & 0.571\stdsize{${\pm}$0.168} \\
          \text{SPAN}-UNet & \textbf{0.885}\stdsize{${\pm}$0.043} & \textbf{0.796}\stdsize{${\pm}$0.069} & \textbf{0.860}\stdsize{${\pm}$0.052} & \textbf{0.757}\stdsize{${\pm}$0.080} & \textbf{0.610}\stdsize{${\pm}$0.042} & \textbf{0.440}\stdsize{${\pm}$0.043} & \textbf{0.783}\stdsize{${\pm}$0.137} & \textbf{0.659}\stdsize{${\pm}$0.173} \\
          \midrule
          \multicolumn{9}{l}{\textbf{Pathology-specific UNI Feature}} \\
          \midrule
          ABMIL & 0.896\stdsize{${\pm}$0.014} & 0.812\stdsize{${\pm}$0.023} & 0.863\stdsize{${\pm}$0.065} & 0.764\stdsize{${\pm}$0.102} & 0.568\stdsize{${\pm}$0.044} & 0.397\stdsize{${\pm}$0.043} &
          0.761\stdsize{${\pm}$0.103} & 0.624\stdsize{${\pm}$0.140} \\
          TransMIL & 0.902\stdsize{${\pm}$0.010} & 0.821\stdsize{${\pm}$0.016} & \underline{0.867}\stdsize{${\pm}$0.068} & \underline{0.770}\stdsize{${\pm}$0.111} & 0.579\stdsize{${\pm}$0.051} & 0.409\stdsize{${\pm}$0.051} & 0.775\stdsize{${\pm}$0.106} & 0.642\stdsize{${\pm}$0.147} \\
          RRT & \underline{0.903}\stdsize{${\pm}$0.002} & \underline{0.822}\stdsize{${\pm}$0.003} & 0.862\stdsize{${\pm}$0.081} & 0.764\stdsize{${\pm}$0.128} & 0.569\stdsize{${\pm}$0.035} & 0.399\stdsize{${\pm}$0.034} & 0.784\stdsize{${\pm}$0.074} & 0.650\stdsize{${\pm}$0.103} \\
          GCN & 0.890\stdsize{${\pm}$0.003} & 0.802\stdsize{${\pm}$0.005} & 0.861\stdsize{${\pm}$0.074} & 0.762\stdsize{${\pm}$0.116} & \underline{0.587}\stdsize{${\pm}$0.054} & \underline{0.417}\stdsize{${\pm}$0.054} & \underline{0.818}\stdsize{${\pm}$0.043} & 0.693\stdsize{${\pm}$0.059} \\
          GAT & 0.890\stdsize{${\pm}$0.005} & 0.802\stdsize{${\pm}$0.009} & 0.864\stdsize{${\pm}$0.071} & 0.766\stdsize{${\pm}$0.112} & 0.580\stdsize{${\pm}$0.061} & 0.410\stdsize{${\pm}$0.061} & 0.817\stdsize{${\pm}$0.066} & \underline{0.694}\stdsize{${\pm}$0.095} \\
          \text{SPAN}-UNet & \textbf{0.908}\stdsize{${\pm}$0.005} & \textbf{0.831}\stdsize{${\pm}$0.008} & \textbf{0.887}\stdsize{${\pm}$0.066} & \textbf{0.802}\stdsize{${\pm}$0.102} & \textbf{0.630}\stdsize{${\pm}$0.033} & \textbf{0.461}\stdsize{${\pm}$0.035} & \textbf{0.830}\stdsize{${\pm}$0.056} & \textbf{0.712}\stdsize{${\pm}$0.081} \\
          \bottomrule
        \end{tabular}
        \begin{tablenotes}
          \footnotesize
          \item[\textdagger] For segmentation tasks, these methods are adapted with corresponding architectures: ABMIL uses MLP, TransMIL uses vanilla Nystromformer, and RRT uses region-based Nystromformer.
        \end{tablenotes}
      \end{threeparttable}
    }
    \vspace{-0.3cm}
  \end{table*}

\section{Experiments}
\label{sec:experiments}

We evaluate \text{SPAN} across multiple classification and segmentation tasks on public datasets using two feature extractors. ResNet50, a long-standing and efficient backbone in WSI analysis that continues to be used for its efficiency in immediate deployment and fast prototyping. UNI~\citep{chen2024towards}, a recent domain-specific foundation model that trades 10× more computation for higher accuracy. For fair comparison, we compare with methods that operate on single-scale inputs, as multi-scale approaches (e.g., HIPT~\citep{chen2022scaling}, H2MIL~\citep{hou2022h}) require extracting and processing features at multiple magnifications, introducing fundamentally different computational and data requirements. Experimental setup details are provided in the Appendix.

\begin{figure}[!ht]
    \vspace{-3mm}
    \centering
    \includegraphics[width=1.0\linewidth]{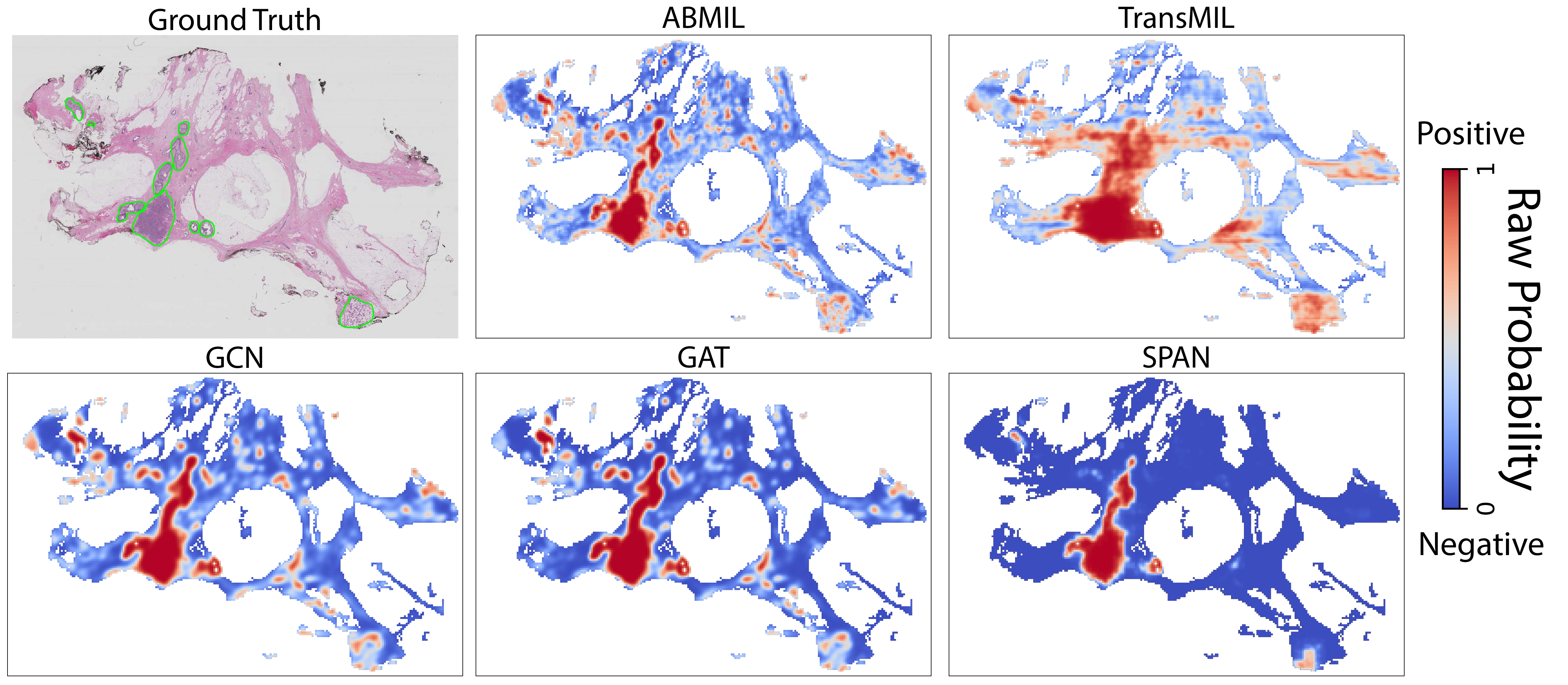}
    \caption{Qualitative comparison of tumor segmentation performance on the unseen test set. The Ground Truth panel depicts the expert-annotated tumor regions enclosed by green contours. The heatmap indicates the predicted probability of tumor presence for each region based on each model’s output in this comparison.}
    \label{fig:viz_seg}
    \vspace{-3mm}
\end{figure}

\noindent\textbf{Overall Performance.}
Tables~\ref{tab:compare_all_datasets} and~\ref{tab:segmentation_by_feature_extractor} show that both \text{SPAN}-MIL and \text{SPAN}-UNet consistently achieve state-of-the-art performance across all tasks, demonstrating superior slide-level and patch-level representation learning capabilities. Notably, for classification, \text{SPAN}-MIL achieves strong performance with only cross-entropy loss, whereas many competing MIL methods rely on additional auxiliary losses and sophisticated training strategies. This simplicity suggests substantial headroom for further improvements through advanced training techniques, while competing approaches that already incorporate multiple losses may face diminishing returns. This success stems from undistorted hierarchical spatial encoding that preserves precise patch relationships, coupled with intrinsic multi-level aggregation for classification and a U-Net-like decoding architecture for segmentation. This architecture allows the model to effectively leverage multi-scale contextual information for precise spatial localization, as illustrated in Fig.~\ref{fig:viz_seg}.

\noindent\textbf{Impact of Feature Extractors.} 
\text{SPAN}'s reliability is further highlighted by its consistent performance gains with pathology-specific UNI features, in contrast to baselines that show inconsistent or degraded results. This suggests that \text{SPAN}'s design becomes more effective when leveraging rich, domain-specific semantic information.

\begin{figure}[ht]
\centering
\includegraphics[width=0.47\textwidth]{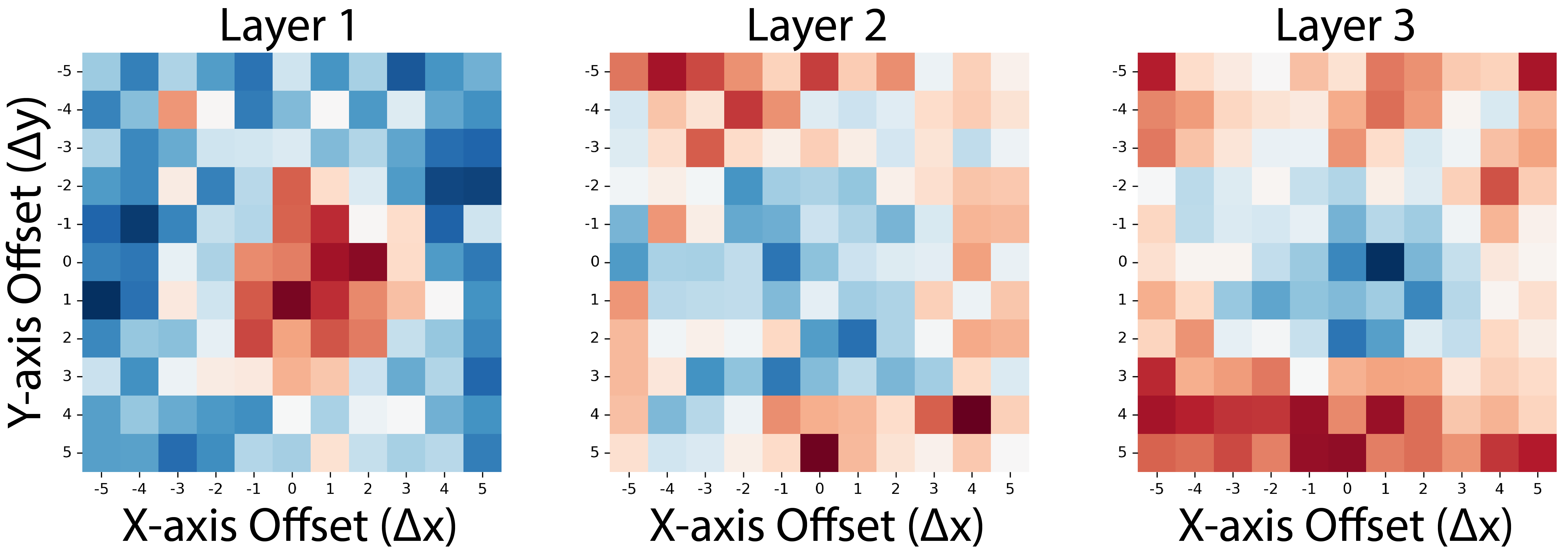}
\caption{Layer-wise visualization of learned RPB in \text{SPAN}. Each heatmap shows attention bias values as a function of relative positional offsets ($\Delta x$, $\Delta y$) between token pairs. Coordinates $(x, y)$ represent the bias when attending to a token at $x$ positions horizontally and $y$ positions vertically relative to the query token. Red and blue indicate higher and lower attention biases, respectively.}
\label{fig:RPB}
\vspace{-3mm}
\end{figure}

\noindent\textbf{Positional Bias Analysis.} 
To understand the internal mechanics of the model, we visualized the learned relative position bias (RPB) in Fig.~\ref{fig:RPB}. The patterns reveal a clear evolution from local attention in the early layers to broad, long-range attention in the deeper layers. This allows \text{SPAN} to dynamically process both fine-grained cellular details and larger tissue architectures across whole-slide images, a flexibility not possible with fixed positional encodings.

\begin{table}[htbp]
    \centering
    \caption{Ablations for different settings.}
    \vspace{-3mm}
    \resizebox{\linewidth}{!}{
        \begin{tabular}{l@{\hspace{0.6em}}cc}
            \toprule
            \multicolumn{3}{l}{\textbf{\text{SPAN}-MIL (Slide-level Representation)}} \\
            \cmidrule(lr){1-3}
            Configuration & Accuracy & AUC \\
            \midrule
            \emph{Attention Pooling} & & \\
            w/o Context Token & $0.893 \pm 0.037$ & $0.931 \pm 0.031$  \\
            w/ Context Token & $0.900 \pm 0.026$ & $0.941 \pm 0.041$  \\
            \midrule
            \emph{Positional Encoding} & & \\
            Axial Alibi & $0.883 \pm 0.039$ & $0.920 \pm 0.029$  \\
            Axial RoPE & $0.880 \pm 0.048$ & $0.917 \pm 0.017$  \\
            None & $0.890 \pm 0.019$ & $0.938 \pm 0.027$  \\
            \midrule
            \emph{Core Modules} & & \\
            No SAC (\(K=S=1\)) & $0.879 \pm 0.037$ & $0.928 \pm 0.026$ \\
            No CAR ($w_{size}=0$) & $0.870 \pm 0.022$ & $0.919 \pm 0.038$ \\
            No Shifted Window & $0.883 \pm 0.039$ & $0.923 \pm 0.049$ \\
            \midrule\midrule
            \multicolumn{3}{l}{\textbf{\text{SPAN}-UNet (Patch-level Representation)}} \\
            \cmidrule(lr){1-3}
            Configuration & Dice & IoU \\
            \midrule
            \emph{Core Modules} & & \\
            No SAC (\(K=S=1\)) & $0.826 \pm 0.059$ & $0.708 \pm 0.091$ \\
            No CAR ($w_{size}=0$) & $0.831 \pm 0.056$ & $0.713 \pm 0.083$ \\
            \midrule
            \emph{Skip Connection Strategy} & & \\
            No Skip Connection & $0.837 \pm 0.059$ & $0.723 \pm 0.088$ \\
            w/ Skip Connection (Add) & $0.848 \pm 0.056$ & $0.739 \pm 0.085$ \\
            \bottomrule
        \end{tabular}
    }
    \label{tab:ablation}
    \vspace{-0.5cm}
\end{table}

\begin{figure}[!ht]
    \vspace{-2mm}
    \centering
    \includegraphics[width=1.0\linewidth]{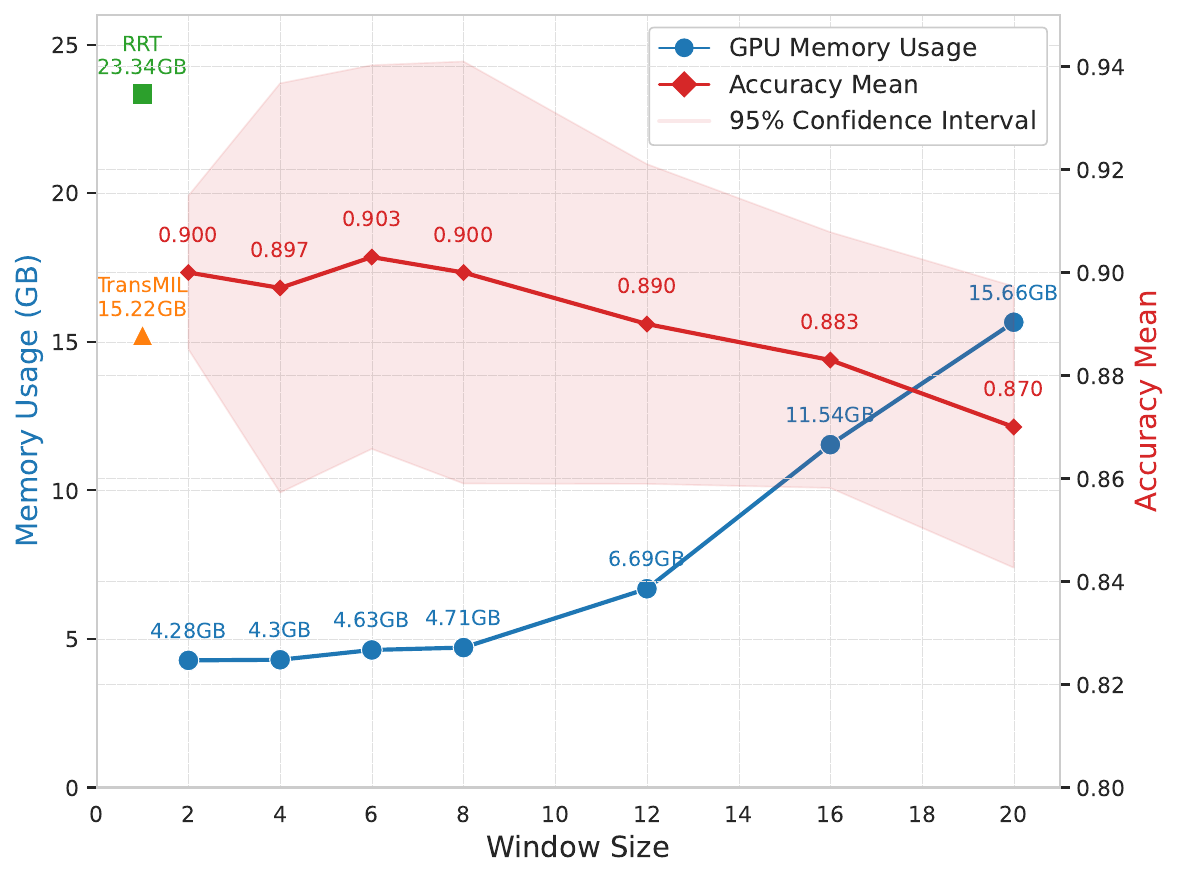}
    \caption{Accuracy and memory usage of \text{SPAN} with window sizes from $2 \times 2$ to $20 \times 20$. Each configuration is evaluated over 5 runs, with the mean accuracy and peak memory usage reported.}
    \label{fig:window_size}
    \vspace{-3mm}
\end{figure}

\noindent\textbf{Ablation Studies.}
We conducted ablation studies on the CAMELYON16 dataset with ResNet50 features to validate the contributions of \text{SPAN}'s components (Table \ref{tab:ablation}, Fig. \ref{fig:window_size}). Aligning with general vision, disabling the SAC's hierarchical downsampling, the CAR's contextual attention, or the shifted-window mechanism all led to significant performance degradation. Interestingly, \text{SPAN} maintains strong performance even without explicit positional encoding. This is because spatial information is inherently encoded through two architectural components: (1) the SAC module's sparse convolutions naturally capture local spatial structures similar to CNNs, and (2) the CAR module's window-based attention implicitly preserves local positional relationships. In this context, RPB serves to further refine and strengthen these spatial relationships. The inferior performance of Axial RoPE and Alibi likely stems from their fixed distance-decay patterns, which conflict with the dynamic spatial relationships that \text{SPAN} learns adaptively across layers (Fig.~\ref{fig:RPB}). For slide-level aggregation, we found that directly using the global context token is effective enough. Finally, as shown in Fig. \ref{fig:window_size}), increasing the window size beyond a certain point does not improve performance in our settings; however, it significantly increases memory usage, which may be attributed to insufficient training data to learn complex feature interactions effectively at larger window sizes.

The results (Table \ref{tab:ablation}) show that our hierarchical pyramid architecture provides a significant performance boost, as the removal of the core SAC or CAR individually resulted in a marked drop in performance. Furthermore, the ablation of skip connections affirms the efficacy of our U-Net-like segmentation design. Removing skip connections resulted in a clear drop in Dice and IoU scores. Collectively, the consistent validation of these diverse, task-specific principles demonstrates the success and flexibility of our framework in bridging the long-standing gap between general deep learning and computational pathology.
\section{Conclusion}
\label{sec:conclusion}

We present \text{SPAN}, a framework that bridges general vision principles and computational pathology. \text{SPAN} advances WSI modeling by (i) learning hierarchical pyramid representations directly from single-scale inputs, (ii) preserving spatial relationships via spatial-adaptive condensation and context-aware refinement, and (iii) supporting flexible variants for classification and segmentation. Extensive experiments confirm that \text{SPAN} delivers consistent gains, establishing it as a WSI backbone that leverages hierarchical and sparsity-aware inductive biases.

\section*{Acknowledgment}
This study is supported by the Department of Defense grant HT9425-23-1-0267.

{
    \small
    \bibliographystyle{ieeenat_fullname}
    \bibliography{main}
}

\clearpage

\setcounter{page}{1}
\renewcommand\thesection{\Alph{section}}
\setcounter{section}{0}
\maketitlesupplementary

\startcontents[appendices]
\section*{Contents of Supplementary Material}
\printcontents[appendices]{}{1}{\normalsize}

\vspace{1em}
\hrule
\vspace{1.5em}

\section{Implementation and Experimental Details}
\subsection{Task-specific Variants}
\label{sec:task_heads}

\subsubsection{\appgreen{SPAN}-MIL: Slide-level Prediction}
We utilize the global context tokens introduced in the CAR module for their comprehensive representations of the WSI across different scales. Let $\mathbf{h}_l^g \in \mathbb{R}^d$ denote the global context token from layer $l \in \{1,\ldots,L\}$. The slide-level representation is computed by:
\begin{equation}
\mathbf{h}^{\text{cls}} = \sum_{l=1}^L \mathbf{h}_l^g.
\end{equation}

The classification prediction is obtained through:
\begin{equation}
\hat{y} = \text{softmax}(W^{\text{cls}}\mathbf{h}^{\text{cls}} + b^{\text{cls}}),
\end{equation}
where $W^{\text{cls}} \in \mathbb{R}^{c \times d}$ and $b^{\text{cls}} \in \mathbb{R}^c$ are learnable parameters, and $c$ is the number of classes.

\subsubsection{\appgreen{SPAN}-UNet: Patch-level Prediction}
SPAN naturally extends to a U-Net~\citep{ronneberger2015u} architecture through its hierarchical sparse design. The decoder maintains architectural symmetry with the encoder, using sparse deconvolution for upsampling in place of the downsampling operations.

Let $\{\mathbf{H}_1, \mathbf{H}_2, \ldots, \mathbf{H}_L\}$ denote the multi-scale feature maps from the encoder, where $\mathbf{H}_l \in \mathbb{R}^{N_l \times d}$ represents features at the $l$-th level.

The decoder generates features $\{\mathbf{G}_1, \mathbf{G}_2, \ldots, \mathbf{G}_L\}$, processed at each stage through:
\begin{equation}
\mathbf{G}_l = \text{SAC}(\text{CAR}(\mathbf{X}_l)) \in \mathbb{R}^{N_l \times d}.
\end{equation}

For the first decoding stage, $\mathbf{X}_1 = \mathbf{H}_L$. For subsequent stages, we implement skip connections by concatenating upsampled features with corresponding encoder features:
\begin{equation}
\mathbf{X}_l = \mathbf{G}_{l-1} \parallel \mathbf{H}_{L - l + 1} \in \mathbb{R}^{N_l \times 2d},
\end{equation}
where $\parallel$ denotes feature concatenation. The final segmentation prediction at position $i$ is:
\begin{equation}
\hat{y}_i = \text{softmax}(W^{\text{seg}} \mathbf{G}_L[i] + b^{\text{seg}}),
\end{equation}
where $W^{\text{seg}} \in \mathbb{R}^{s \times d}$ and $b^{\text{seg}} \in \mathbb{R}^s$ are learnable parameters, and $s$ is the number of segmentation classes.

\subsection{Experimental Setup}
\label{apx:exp}
\subsubsection{Classification Datasets}
WSI classification involves automatically categorizing tissues based on histopathological features, an essential process for accurate diagnosis, grading, and personalized treatment planning. We assessed SPAN's classification performance on three distinct diagnostic tasks, specifically tumor detection using the CAMELYON16 dataset~\citep{bejnordi2017diagnostic}, tumor grading employing the BRACS dataset~\citep{brancati2022bracs}, and HER2 biomarker status prediction using the Yale-HER2 dataset~\cite{farahmand2022deep}.

We followed the same strategy as above: all available slides were pooled, randomly shuffled, and split into training ($\sim$70\%), validation ($\sim$15\%), and test ($\sim$15\%). Experiments were repeated under five random seeds (0--4). All models were trained using cross-entropy loss. Model selection is based on validation set performance. Crucially, final predictions are made via direct class probability argmax, without any post-hoc threshold optimization, to better mirror real-world clinical deployment scenarios.

\subsubsection{Segmentation Datasets} Slide-level segmentation requires precise pixel-level delineation of tumor regions, a challenging task crucial for diagnosis and prognosis. To rigorously evaluate SPAN's performance, we used fully annotated slides from multiple datasets: SegCAMELYON, Yale-HER2~\citep{farahmand2022deep}, and BACH~\citep{aresta2019bach}. To construct the SegCAMELYON benchmark, we curated tumor-positive slides from CAMELYON16~\citep{bejnordi2017diagnostic} and CAMELYON17~\citep{bandi2018detection}, applied exclusion masks to remove ambiguous regions, and consolidated the processed samples into a unified dataset.

All available slides were pooled, randomly shuffled, and split into training ($\sim$70\%), validation ($\sim$10\%), and test ($\sim$20\%). Experiments were repeated under five random seeds (0--4) to ensure robustness. Patches with over 20\% tumor area are labeled positive for patch-level ground truth generation. For segmentation, we adopted 3-layer GCN and GAT models with 8-adjacent connectivity, following standard WSI analysis practices~\citep{hou2022h,chen2021whole,wu2023graph}. Model selection is based on validation set performance. Crucially, final predictions are made via direct class probability argmax, without any post-hoc threshold optimization, to better mirror real-world clinical deployment scenarios.

For segmentation training, we employed a hybrid loss that combines cross-entropy (CE) and Dice loss. Specifically, given the predicted probability map $\mathbf{p}$ and the ground-truth mask $\mathbf{y}$, we compute the standard pixel-wise CE loss $\mathcal{L}_{\text{CE}}(\mathbf{p}, \mathbf{y})$ and the Dice loss $\mathcal{L}_{\text{Dice}}(\mathbf{p}, \mathbf{y})$. The final objective is defined as:
\[
\mathcal{L} =
\begin{cases}
(1 - \lambda)\,\mathcal{L}_{\text{CE}} + \lambda\,\mathcal{L}_{\text{Dice}}, & \text{if } \sum \mathbf{y} > 0, \\
\mathcal{L}_{\text{CE}}, & \text{otherwise},
\end{cases}
\]
where $\lambda=0.75$ is the Dice weight. This design follows common practices in computer vision community, encouraging accurate boundary delineation when positives are present. All baseline methods were trained under this unified loss function for fair comparison.

\subsubsection{Slide Preprocessing} Our preprocessing pipeline extends CLAM~\citep{lu2021data} by adding a grid alignment step, adjusting patch boundaries to the nearest multiple of 224 pixels for precise spatial coordinates.

To evaluate feature-space adaptability, we used two pre-trained encoders to generate patch-level features from all datasets at 20x magnification. All patches were resized to 224$\times$224 pixels prior to feature extraction. Our preprocessing pipeline addresses coordinate inconsistencies that arise from CLAM's background filtering mechanism. The original CLAM pipeline can generate patches with irregular starting coordinates due to tissue contour boundaries, making it difficult to establish consistent spatial relationships in a regular grid system. To resolve this, we introduced a grid alignment step that extends tissue contours to align with 224$\times$224 pixel boundaries before patch extraction.

\begin{algorithm}[ht]
    \caption{Expand Contours}
    \label{algo:grid-alignment}
    \SetKwProg{Fn}{def}{:}{\KwRet}
    \KwSty{global} step\_size = 224 \\
    \Fn{extend\_contour(start\_x, start\_y, w, h)}{
        w += start\_x \% step\_size \\
        h += start\_y \% step\_size \\
        start\_x -= start\_x \% step\_size \\
        start\_y -= start\_y \% step\_size \\
        \KwSty{return} start\_x, start\_y, w, h
    }
    \comment{contour: (start\_x, start\_y, w, h)} \\
    contour = extend\_contour(contour)
\end{algorithm}

This alignment ensures that all patches map precisely to a regular grid coordinate system, eliminating potential rounding errors in spatial relationship modeling. 

\subsubsection{Patch Feature Extractor}

In all experiments, the weights of these encoders were kept frozen to ensure a consistent feature extraction process.

\paragraph{ResNet50} As a standard baseline, we used a ResNet50 model pre-trained on ImageNet~\citep{he2016deep}. Following common practice in WSI analysis, we removed the final fully connected classification layer and used the output of the global average pooling layer. This process yields a 1024-dimensional feature vector for each patch, representing general-purpose visual features learned from natural images.

\paragraph{UNI} We utilized UNI~\citep{chen2024towards}, a foundation model specifically tailored for computational pathology. Unlike general-purpose vision models, UNI is built upon a ViT-Large architecture and pretrained via self-supervised learning on a massive pan-cancer dataset comprising over 100 million tissue patches from more than 100,000 WSIs. This extensive domain-specific exposure enables the model to capture subtle histological patterns and high-level tissue semantics that are often missed by ImageNet supervised baselines.

\begin{algorithm}[ht]
      \caption{SPAN Backbone with Rulebook Mechanism}
      \label{algo:span-backbone}
      \SetKwProg{Fn}{def}{:}{\KwRet}
      \KwIn{$\mathbf{P} \in \mathbb{N}^{N \times 2}$
  (coordinates), $\mathbf{X} \in \mathbb{R}^{N \times d}$
  (features)}
      \KwOut{Refined features and global context}

      \For{each layer in backbone}{
          \tcp{SAC Module: Sparse Convolution Rulebook}
          $\mathbf{P}_{\text{out}} \leftarrow$
  compute\_output\_coords($\mathbf{P}$, $K$, $S$, $D$) \\
          $\mathcal{R}_{\text{sparse}} \leftarrow$
  build\_sparse\_rulebook($\mathbf{P}$,
  $\mathbf{P}_{\text{out}}$, $\mathcal{K}$) \\
          $\mathbf{X} \leftarrow$
  execute\_sparse\_conv($\mathbf{X}$,
  $\mathcal{R}_{\text{sparse}}$, $\mathbf{W}$) \\

          \tcp{CAR Module: Sparse Attention Rulebook}
          $\mathcal{W} \leftarrow$
  generate\_windows($\mathbf{P}_{\text{out}}$, window\_size) \\
          $\mathcal{R}_{\text{local}} \leftarrow \{(i,j) \mid
  i,j \in w, \forall w \in \mathcal{W}\}$ \\
          $\mathcal{R}_{\text{global}} \leftarrow \{(i, N+1),
  (N+1, i) \mid i \in [1,N]\}$ \\
          $\mathbf{X} \leftarrow$
  execute\_attention($\mathbf{X}$, $\mathcal{R}_{\text{local}}$,
   $\mathcal{R}_{\text{global}}$) \\

          $\mathbf{P} \leftarrow \mathbf{P}_{\text{out}}$ \\
      }
      \KwRet{$\mathbf{X}$, global\_token}
  \end{algorithm}

  \begin{algorithm}[ht]
      \caption{Build Sparse Attention Rulebook}
      \label{algo:attention-rulebook}
      \SetKwProg{Fn}{def}{:}{\KwRet}
      \KwIn{$\mathbf{P} \in \mathbb{N}^{N \times 2}$
  (coordinates), $w$ (window size)}
      \KwOut{$\mathcal{R}_{\text{local}},
  \mathcal{R}_{\text{global}}$ (attention rulebooks)}

      \tcp{Create coordinate hash mapping}
      hash\_ids $\leftarrow$ arange(1, $N+1$) \\
      coord\_transpose $\leftarrow \mathbf{P}$.transpose() \\
      spatial\_bounds $\leftarrow$ (max(coord\_transpose[0]) +
  1, max(coord\_transpose[1]) + 1) \\
      coord\_tensor $\leftarrow$
  create\_sparse\_coo(coord\_transpose, hash\_ids,
  spatial\_bounds) \\
      index\_matrix $\leftarrow$ coord\_tensor.to\_dense() \\

      \tcp{Generate attention windows via spatial indexing}
      \eIf{index\_matrix.size() $< 2w \times 2w$}{
          \tcp{Compact space: full attention}
          spatial\_indices $\leftarrow$ arange(num\_elements) \\
          query\_idx $\leftarrow$
  spatial\_indices.repeat\_interleave(num\_elements) \\
          key\_idx $\leftarrow$
  spatial\_indices.repeat(num\_elements) \\
      }{
          \tcp{Extended space: windowed attention}
          window\_blocks $\leftarrow$
  generate\_windows(index\_matrix, $w$, mode) \\
          block\_capacity $\leftarrow (2w)^2$ \\
          intra\_indices $\leftarrow$ arange(block\_capacity) \\
          query\_idx $\leftarrow$
  intra\_indices.unsqueeze(1).repeat(1,
  block\_capacity).flatten() \\
          key\_idx $\leftarrow$
  intra\_indices.repeat(block\_capacity) \\

          query\_hash $\leftarrow$
  window\_blocks.flatten()[query\_idx] \\
          key\_hash $\leftarrow$
  window\_blocks.flatten()[key\_idx] \\
      }

      \tcp{Filter valid mappings and normalize hash indices}
      valid\_mask $\leftarrow$ (query\_hash $\neq$ 0) $\land$
  (key\_hash $\neq$ 0) $\land$ (query\_hash $\neq$ key\_hash) \\
      $\mathcal{R}_{\text{local}} \leftarrow$
  (query\_hash[valid\_mask] - 1, key\_hash[valid\_mask] - 1) \\

      \tcp{Global context rulebook}
      $\mathcal{R}_{\text{global}} \leftarrow \{(\alpha,
  N+\beta), (N+\beta, \alpha) \mid \alpha \in [0, N-1], \beta
  \in [0, \text{num\_ctx}-1]\}$ \\

      \KwRet{$\mathcal{R}_{\text{local}},
  \mathcal{R}_{\text{global}}$}
  \end{algorithm}

  \begin{algorithm}[ht]
      \caption{Spatial Window Indexing}
      \label{algo:spatial-windowing}
      \SetKwProg{Fn}{def}{:}{\KwRet}
      \KwIn{index\_matrix, $w$ (window radius), mode}
      \KwOut{Active window blocks}

      $h, width \leftarrow$ index\_matrix.size() \\

      \tcp{Compute spatial alignment padding}
      row\_align $\leftarrow (2w - h \bmod 2w) \bmod 2w$ \\
      col\_align $\leftarrow (2w - width \bmod 2w) \bmod 2w$ \\

      \If{row\_align $> 0$ or col\_align $> 0$}{
          index\_matrix $\leftarrow$ spatial\_pad(index\_matrix,
   alignment\_spec, mode) \\
      }

      \tcp{Efficient spatial tessellation}
      window\_tessellation $\leftarrow$ index\_matrix.unfold(0,
  $2w$, $2w$).unfold(1, $2w$, $2w$) \\

      \tcp{Filter active windows by occupancy}
      occupancy\_map $\leftarrow$
  window\_tessellation.sum(dim=[-2, -1]) \\
      \KwRet{window\_tessellation[occupancy\_map $> 0$]}
  \end{algorithm}

  \begin{algorithm}[ht]
      \caption{Execute Rulebook-based Attention}
      \label{algo:rulebook-attention}
      \SetKwProg{Fn}{def}{:}{\KwRet}
      \KwIn{$\mathbf{Q}, \mathbf{K}, \mathbf{V}$ (projections),
  $\mathcal{R}_{\text{local}}, \mathcal{R}_{\text{global}}$
  (rulebooks)}
      \KwOut{$\mathbf{H}_{\text{out}}$ (refined features)}

      \tcp{Local attention via spatial rulebook}
      \For{$(\alpha, \beta) \in \mathcal{R}_{\text{local}}$}{
          $\phi_{\alpha\beta} \leftarrow
  \frac{\mathbf{q}_\alpha^\top \mathbf{k}_\beta}{\sqrt{d}} +
  \mathcal{B}(\mathbf{P}[\alpha] - \mathbf{P}[\beta])$ \\
      }
      $\mathbf{H}_{\text{local}} \leftarrow$
  apply\_rulebook\_softmax($\{\phi_{\alpha\beta}\}$,
  $\mathbf{V}$, $\mathcal{R}_{\text{local}}$) \\

      \tcp{Global attention via context rulebook}
      \For{$(\alpha, \beta) \in \mathcal{R}_{\text{global}}$}{
          $\psi_{\alpha\beta} \leftarrow
  \frac{\mathbf{q}_\alpha^\top \mathbf{k}_\beta}{\sqrt{d}}$ \\
      }
      $\mathbf{H}_{\text{global}} \leftarrow$
  apply\_rulebook\_softmax($\{\psi_{\alpha\beta}\}$,
  $\mathbf{V}$, $\mathcal{R}_{\text{global}}$) \\

      $\mathbf{H}_{\text{out}} \leftarrow
  \mathbf{H}_{\text{local}} + \mathbf{H}_{\text{global}}$ \\
      \KwRet{$\mathbf{H}_{\text{out}}$}
  \end{algorithm}

\section{Runtime}
In the CAMELYON16 dataset, SPAN-MIL training runs 12.32 seconds per slide, compared to 3.09 seconds for ABMIL and 16.96 seconds for TransMIL.

\section{Experiments on Clinical and Biological Tasks}

In addition to the human-annotated computer vision benchmarks presented in the main paper, we further evaluate SPAN on survival prediction, a clinical task that uses patient outcome supervision. Unlike traditional computer vision tasks where ground truth is defined by pathologists' visual perception, this task relies on objective biological signals from other modalities, including patient survival outcomes, which may not have obvious visual correlates. Consequently, it requires the model to discover complex, non-trivial morphological patterns that are often subtle or invisible to the human eye. These experiments are complementary to the results in the main paper and demonstrate the applicability of SPAN beyond conventional vision benchmarks. We conducted 5 independent runs with different random seeds (0–4) using UNI features, where each run randomly splits the data into training/validation/test sets. For each run, we select the checkpoint that performs best on the validation set and report its corresponding test performance. We then report the mean and standard deviation across the 5 runs.

\subsection{Survival Analysis}

Survival prediction is a fundamental task in oncology that aims to estimate the risk of adverse events such as death or recurrence for each patient. Accurate risk stratification is crucial for personalized treatment planning. We evaluated SPAN for patient survival prediction on three TCGA cohorts: LGG (Lower-Grade Glioma), LUAD (Lung Adenocarcinoma), and LUSC (Lung Squamous Cell Carcinoma)~\cite{cancer2014comprehensive}.

For each cohort, we extracted clinical survival information including overall survival time and vital status. To prevent data leakage, we retained only one slide per patient by excluding duplicates based on Patient ID. We discretized the continuous survival times into $K=3$ risk groups using quantile-based binning on uncensored patients, and then applied the resulting bins to all samples. Slides without valid survival annotations or with zero survival time were excluded from the analysis. For fair comparison with baseline methods, we adopted the same data split protocol: one-third of the data was reserved as the test set, and 15\% of the remaining two-thirds was used for validation, with the rest allocated to training.

We trained the models using the negative log-likelihood survival loss~\citep{kvamme2019time}, which accounts for both censored and uncensored events. The loss function is defined as:
\begin{equation}
\mathcal{L} = -\frac{1}{N}\sum_{i=1}^N \left[ (1-c_i)(\log h_{y_i} + \log S_{y_i}) + c_i \log S_{y_i+1} \right],
\end{equation}
where $h$ represents the predicted discrete hazards, $S$ is the survival probability, $y_i$ is the discretized survival label, and $c_i$ indicates censorship status (1 for censored, 0 for event observed).

Table~\ref{tab:survival_all} summarizes the results. SPAN consistently outperforms state-of-the-art MIL baselines across all three datasets. We attribute the sub-random performance of several baselines to our rigorous evaluation protocol, specifically the strict validation-based model selection and the discrete survival objective (using $K=3$ risk groups). These constraints expose the tendency of standard MIL methods to overfit on limited data, whereas SPAN's hierarchical structure promotes robust performance.

\begin{table}[ht]
\centering
\small
\caption{Survival prediction performance using UNI features. Baseline results are compared with SPAN-MIL. Values are mean C-index $\pm$ standard deviation.}
\label{tab:survival_all}

\resizebox{0.95\linewidth}{!}{
\begin{tabular}{lccc}
\toprule
\textbf{Method} & \textbf{LGG} & \textbf{LUAD} & \textbf{LUSC} \\
\midrule
ACMIL      & $0.438 \pm 0.092$ & $0.460 \pm 0.059$ & $0.500 \pm 0.048$ \\
ABMIL     & $0.416 \pm 0.101$ & $0.452 \pm 0.070$ & $0.500 \pm 0.043$ \\
CLAM-MB   & $0.394 \pm 0.088$ & $0.455 \pm 0.064$ & $0.519 \pm 0.046$ \\
CLAM-SB   & $0.436 \pm 0.094$ & $0.447 \pm 0.074$ & $0.528 \pm 0.027$ \\
DSMIL      & $0.411 \pm 0.104$ & $0.471 \pm 0.028$ & $0.498 \pm 0.081$ \\
RRTMIL     & $0.407 \pm 0.094$ & $0.476 \pm 0.043$ & $0.455 \pm 0.049$ \\
TransMIL   & $0.419 \pm 0.072$ & $0.486 \pm 0.030$ & $0.488 \pm 0.068$ \\
\midrule
\textbf{SPAN-MIL} & 
\textbf{$0.647 \pm 0.034$} &
\textbf{$0.570 \pm 0.044$} &
\textbf{$0.584 \pm 0.046$} \\
\bottomrule
\end{tabular}
}
\end{table}

\section{Potential Applications and Limitations}

We believe SPAN provides a meaningful contribution to the digital pathology community. By adapting hierarchical vision architectures to the sparse and irregular structure of WSIs, SPAN establishes a robust and flexible computational foundation that aligns more closely with the intrinsic geometry of gigapixel pathology images. Beyond the benchmarks presented in this work, the framework naturally opens pathways for a broad range of advanced modeling strategies and clinical applications.

\subsection{Foundation for Advanced Training Strategies}
Although SPAN already achieves strong performance using a purely supervised objective, the architecture is well positioned to benefit from more sophisticated training schemes. Similar to how ABMIL has served as a general-purpose backbone for methods such as CLAM and ACMIL, SPAN can function as a versatile MIL foundation. Strategies such as knowledge distillation, curriculum-based hard negative mining, or task-specific auxiliary losses could be incorporated without altering the core design. Because SPAN preserves local spatial context and maintains stable hierarchical representations, these techniques may further enhance performance on challenging or fine-grained clinical tasks.

\subsection{Large-scale Slide-level Pretraining}
The hierarchical sparse design of SPAN is inherently well suited for large-scale whole-slide pretraining. Unlike patch-level pretraining paradigms that focus on isolated ROI features, SPAN preserves multi-resolution context and global tissue structure, making it compatible with emerging slide-level foundation model training~\cite{ding2025multimodal,xu2024whole,chen2025slidechat}. By masking or perturbing regions across the slide while retaining SPAN's spatial hierarchy, the model could learn robust and generalizable representations from large collections of unlabeled WSIs. Coupling SPAN with pathology reports or synthetic captions further enables slide–text alignment, similar to recent whole-slide vision and language models. Such pretraining strategies may substantially improve downstream performance, particularly for rare clinical conditions or limited-data settings where strong slide-level representations are essential.

\subsection{Efficient Patch-level Extractor Adaptation}
Although SPAN is currently trained with frozen patch-level features, the framework naturally supports parameter-efficient fine-tuning of the underlying foundation model. By inserting LoRA adapters~\cite{hu2022lora} into a patch-level backbone such as UNI, one can selectively update the feature extractor while keeping the SPAN hierarchy fixed. This enables end-to-end optimization across both modules with minimal computational overhead. It provides a practical path for adapting large vision foundation models to domain-specific clinical tasks without the cost of full fine-tuning.

\subsection{Complex Multimodal Tasks}
Our results indicate that SPAN effectively captures both fine-grained spatial details and broader contextual relationships. This makes it a promising backbone for future vision and language modeling in computational pathology. Tasks such as report generation, captioning, or visual question answering require accurate grounding of visual features~\cite{chen2025slidechat, chen2024wsicaption, chen2024wsi, chen2025segment}, an area where current patch-based encoders often struggle due to limited positional structure. The spatially coherent representations produced by SPAN may therefore offer distinct advantages for multimodal reasoning over whole-slide images.

\subsection{Limitations}
Our work primarily establishes SPAN as a supervised learning baseline. We have not yet explored its integration with self-supervised slide-level pretraining, multimodal foundation models, or domain adaptation frameworks, all of which may further expand the model's utility. In addition, the current experiments only use single-scale inputs. Extending SPAN from single-scale to multi-scale inputs is a natural next step, and its hierarchical design may enable a more coherent integration of different magnifications than current isotropic multi-scale methods requiring multi-scale inputs, such as HIPT~\citep{chen2022scaling}, H2MIL~\citep{hou2022h}, and ZoomMIL~\citep{thandiackal2022differentiable}. Also, although the rulebook-based implementation is efficient, further optimization for specialized hardware accelerators such as GPUs with sparse kernels could improve speed for clinical deployment. Finally, our evaluation is limited to publicly available datasets. Assessing robustness across institutions, scanners, and staining protocols remains an important direction for future work.

\end{document}